\documentclass[sigconf,natbib,screen=true]{acmart}

\PassOptionsToPackage{dvipsnames}{xcolor}

\usepackage{dsfont}
\usepackage{acronym}
\usepackage{amsmath}
\usepackage[noend]{algorithmic}
\usepackage{algorithm}
\usepackage{bbm}
\usepackage{beramono}
\usepackage{bm}
\usepackage{booktabs}
\usepackage[skip=0pt]{caption}
\usepackage{cleveref}
\usepackage[T1]{fontenc}
\usepackage{graphicx}
\usepackage{hyperref}
\usepackage{listings}
\usepackage{multirow}
\usepackage{natbib}
\usepackage{tabularx}
\usepackage[dvipsnames]{xcolor}
\usepackage{enumerate}
\usepackage[inline]{enumitem}
\usepackage{numprint}
\usepackage[normalem]{ulem}
\usepackage{mleftright}
\usepackage{balance}
\usepackage{makecell}
\usepackage{tabularx}

\copyrightyear{2022}
\acmYear{2022}
\setcopyright{rightsretained}

\acmConference[SIGIR '22]{Proceedings of the 45th International ACM SIGIR Conference on Research and Development in Information Retrieval}{July 11--15, 2022}{Madrid, Spain.}
\acmBooktitle{Proceedings of the 45th Int'l ACM SIGIR Conference on Research and Development in Information Retrieval (SIGIR '22), July 11--15, 2022, Madrid, Spain}
\acmDOI{10.1145/3477495.3531842}
\acmISBN{978-1-4503-8732-3/22/07}

\usepackage{etoolbox}
\makeatletter
\patchcmd{\maketitle}{\@copyrightpermission}{
  \begin{minipage}{0.3\columnwidth}
    \href{http://creativecommons.org/licenses/by/4.0/}{\includegraphics[width=0.90\textwidth]{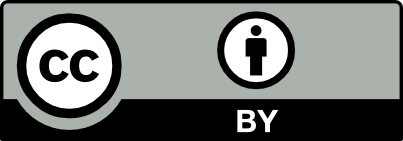}}
  \end{minipage}\hfill
  \begin{minipage}{0.7\columnwidth}
    \href{http://creativecommons.org/licenses/by/4.0/}{This work is licensed under a Creative Commons Attribution International 4.0 License.}
    \end{minipage}
    
   \vspace{5pt}
}{}{}

\makeatother


\definecolor{rj}{RGB}{0, 150, 0}
\definecolor{mdr}{RGB}{200, 0, 0}
\definecolor{ho}{RGB}{0, 50, 150}

\acrodef{IR}{Information Retrieval}
\acrodef{LTR}{Learning-to-Rank}
\acrodef{ARP}{Average Relevance Position}
\acrodef{DCG}{Discounted Cumulative Gain}
\acrodef{EM}{Expectation Maximization}
\acrodef{CTR}{Click-Through-Rate}
\acrodef{IPS}{Inverse-Propensity-Scoring}
\acrodef{LogOpt}{Logging-Policy Optimization Algorithm}
\acrodef{RCTR}{Relevant-Click-Through-Rate}
\acrodef{DBGD}{Dueling Bandit Gradient Descent}
\acrodef{COLTR}{Counterfactual Online Learning to Rank}
\acrodef{PDGD}{Pairwise Differentiable Gradient Descent}
\acrodef{NRCTR}{Normalized RCTR}
\acrodef{NDCG}{Normalized DCG}

\acrodef{PL}{Plackett-Luce}

\newcolumntype{Y}{>{\centering\arraybackslash}X}

\allowdisplaybreaks

\author{Harrie Oosterhuis}
\affiliation{%
	\institution{Radboud University}
	\city{Nijmegen}
	\country{The Netherlands}
}
\email{harrie.oosterhuis@ru.nl}

\acmSubmissionID{1493}

\title{Learning-to-Rank at the Speed of Sampling: Plackett-Luce Gradient Estimation With Minimal Computational Complexity}

\allowdisplaybreaks

\begin{document}

\begin{abstract}
Plackett-Luce gradient estimation enables the optimization of stochastic ranking models within feasible time constraints through sampling techniques.
Unfortunately, the computational complexity of existing methods does not scale well with the length of the rankings, i.e.\ the ranking cutoff, nor with the item collection size.

In this paper, we introduce the novel PL-Rank-3 algorithm that performs unbiased gradient estimation with a computational complexity comparable to the best sorting algorithms.
As a result, our novel learning-to-rank method is applicable in any scenario where standard sorting is feasible in reasonable time.
Our experimental results indicate large gains in the time required for optimization, without any loss in performance.
For the field, our contribution could potentially allow state-of-the-art learning-to-rank methods to be applied to much larger scales than previously feasible.
\end{abstract}

%\fancyhead{}

\maketitle

\acresetall

\section{Introduction}
\label{sec:intro}

\ac{LTR} methods optimize ranking systems for search and recommendation purposes~\citep{liu2009learning}.
In the field of \ac{IR}, they are generally deployed to maximize well-known ranking metrics such as: \ac{DCG} or precision~\citep{jarvelin2002cumulated}.
The main difficulty with the \ac{LTR} task is that ranking metrics are non-differentiable, discrete and very non-smooth due to the underlying sorting process~\citep{burges2010ranknet}.
In broad terms, the existing solutions to this problem can be divided into two categories:
heuristic bounds or approximations of ranking metrics or their gradients~\citep{fuhr1989optimum, joachims2002optimizing, taylor2008softrank, bruch2019revisiting, wang2018lambdaloss, burges2010ranknet};
and optimizing a probabilistic ranking model instead of a deterministic model~\citep{xia2008listwise, cao2007learning, oosterhuis2021computationally, ustimenko2020stochasticrank}.
The latter category does not have issues with differentiability or smoothness because their methods optimize a probabilistic distribution over rankings.
Recently, probabilistic ranking models have also received additional attention for their applicability to ranking fairness tasks~\citep{diazevaluatingstochastic}.
Since probabilistic models are able to divide exposure over items more fairly than deterministic models.

\citet{oosterhuis2021computationally} introduced the PL-Rank method to efficiently optimize \ac{PL} ranking models: a specific type of ranking model based on decision theory~\citep{plackett1975analysis, luce2012individual}.
They use Gumbel sampling~\citep{gumbel1954statistical, bruch2020stochastic} to quickly sample many rankings from a \ac{PL} ranking model, and subsequently, apply the PL-Rank-1 or PL-Rank-2 algorithm to these samples to unbiasedly approximate the gradient of a ranking metric w.r.t.\ the model.
While PL-Rank provides a significant contribution to the \ac{LTR} field, we recognize two shortcomings in the work of \citet{oosterhuis2021computationally}:
neither PL-Rank-1 or PL-Rank-2 scale well with long rankings;
and they do not compare PL-Rank to the earlier and comparable StochasticRank algorithm~\citep{ustimenko2020stochasticrank}.

This paper addresses both of these shortcomings: our main contribution is the novel PL-Rank-3 algorithm that computes the same approximation as PL-Rank-2 while minimizing its computational complexity.
PL-Rank-3 has computional costs in the same order as the best sorting algorithms; we posit that this is the lowest order possible for a metric-based \ac{LTR} method.
Our experimental comparison includes both the previous PL-Rank-2 and StochasticRank as baselines, providing the first comparison between these two methods.
The introduction of PL-Rank-3 is exciting for the \ac{LTR} field, as it pushes the limit of the minimal computational complexity that \ac{LTR} methods can have.
Potentially, it may enable future \ac{LTR} to be applied to much larger scales than currently feasible.

\begin{table*}[t]
\captionsetup{singlelinecheck=off}
\caption{
Overview of \ac{LTR} methods in terms of their relevant theoretical properties:
(i) Reliance on an iteration over item pairs;
(ii) Directed by the model's full-ranking behavior;
(iii) Directed by metric to be optimized;
(iv) Usage of sample-based approximation;
(v) Applicability to general rank-based exposure metrics (e.g.\ for ranking fairness);
(vi) Computational complexity w.r.t.\ number of items $D$ and length of ranking $K$.
The properties of a method are indicated by \checkmark or ? when it is open to interpretation and
$D^{2*}$ indicates the number of item-pairs with unequal relevance labels.
}
\label{table:properties}
\begin{tabularx}{\textwidth}{X  c  c  c  c  c  c  c}
\toprule
\multicolumn{1}{ c |}{\multirow{2}{2cm}{\it method name}}
& \multicolumn{1}{ c |}{\multirow{2}{1.1cm}{\it pairwise}}
& \multicolumn{1}{ c |}{\it ranking-}
& \multicolumn{1}{ c |}{\it metric-}
& \multicolumn{1}{ c |}{\it sample}
& \multicolumn{1}{ c |}{\it rank-based}
& \multicolumn{1}{ c |}{\it computational}
& \multicolumn{1}{ c }{ \multirow{2}{1cm}{\it notes}}
\\
\multicolumn{1}{ c |}{}
& \multicolumn{1}{ c |}{}
& \multicolumn{1}{ c |}{\it based}
& \multicolumn{1}{ c |}{\it based}
& \multicolumn{1}{ c |}{\it approximation}
& \multicolumn{1}{ c |}{\it exposure}
& \multicolumn{1}{ c |}{\it complexity}
\\
\midrule
Pointwise~\citep{fuhr1989optimum} & &&&&& $D$ & not an \ac{LTR} loss
\\
SoftMax~Cross-Entropy~\citep{qin2021neural} &  & &&&& $D$ & 
\\
Pairwise~\citep{joachims2002optimizing} & \checkmark &&&&& $D^2$* & memory efficient 
\\
Listwise/ListMLE~\citep{xia2008listwise,cao2007learning} &  & \checkmark&&&& $DK$ & \hspace{3cm}
\\
SoftRank~\citep{taylor2008softrank} & \checkmark & ? & \checkmark & & & $D^3$ & 
\\
ApproxNDCG~\citep{bruch2019revisiting} & \checkmark & ? & \checkmark & & & $D^2$ & proven bound
\\
LambdaRank/Loss~\citep{wang2018lambdaloss, burges2010ranknet} & \checkmark & \checkmark&\checkmark&&& $D^{2*}  + D\log(D)$ & proven bound 
\\
StochasticRank~\citep{ustimenko2020stochasticrank} & ? & \checkmark & \checkmark & \checkmark & \checkmark & $DK$ & policy-gradient
\\
PL-Rank-1/2~\citep{oosterhuis2021computationally} &  & \checkmark & \checkmark & \checkmark & \checkmark & $DK$ & policy-gradient
\\
PL-Rank-3~(ours) &  & \checkmark & \checkmark & \checkmark & \checkmark & $D + K\log(D)$ & policy-gradient
\\
\bottomrule
\end{tabularx}
\end{table*}

\section{Background: PL-Rank-2}
Let $\pi$ indicate a \ac{PL} ranking model based on a scoring function $f$ with $f(d)$ indicating the score for item $d$; the probability of sampling ranking $y$ from item set $D$ from $\pi$ is then:
\begin{equation*}
\pi(y) = \prod \pi(d |\, y_{1:k-1}, D),
\,
 \pi(d |\, y_{1:k-1}, D) = \frac{e^{f(d)} \mathds{1}[d \not\in y_{1:k-1}]}{\sum_{d' \in D \setminus y_{1:k-1}} e^{f(d')}},
\end{equation*}
where $y_{1:k-1}$ indicates the ranking from rank $1$ up to $k-1$. 
In other words, the probability of placing $d$ at rank $k$ is $e^{f(d)}$ divided by the $e^{f(d')}$ of all unplaced items (similar to a SoftMax activation function), unless $d$ was already placed at an earlier rank then it has a zero probability.
The probability of the entire ranking $y$ is simply the product of all its individual item placements.

PL-Rank optimizes $f$ so that the metric value of sampled rankings are maximized in expectation~\citep{oosterhuis2021computationally}.
PL-Rank assumes ranking metrics can be decomposed into weights per rank $\theta_k$ and relevance of an item $\rho_{q,d}$~\citep{moffat2017incorporating}.
Its objective is thus to maximize the expected metric value over its sampling procedure,
for a single query $q$:
\begin{equation*}
\mathcal{R}(q) = \sum_y \pi(y) \sum_{d \in D_q} \theta_{\text{rank(f,q,d)}} \rho_{q,d} = \mathbb{E}_y\bigg[ \sum_{d \in D_q} \theta_{\text{rank(f,q,d)}} \rho_{q,d} \bigg].
\end{equation*}
The rank weights can be chosen to match well-known ranking metrics, e.g.\ DCG:
$\theta_k^\text{DCG@K} = \mathds{1}[k \leq K]/\log_2(1+k)$~\citep{jarvelin2002cumulated}; or precision:
$\theta_k^\text{prec} = \mathds{1}[k \leq K]$.
To keep our notation brief, we will omit $q$ when denoting the relevances $\rho_{q,d} = \rho_{d}$.
\citet{oosterhuis2021computationally} based the PL-Rank-2 method on the following formulation of the policy gradient:
\begin{equation}
\begin{split}
\frac{\delta \mathcal{R}(q)}{\delta f(d)} 
&=
\mathbb{E}_{y} \Bigg[
\Bigg(\sum_{k=\text{rank}(d, y)+1}^K
 \theta_k
\rho_{y_k}
\Bigg)
\\
&\quad\,\,\,\,\,\,\;\;
+
\sum_{k=1}^{\text{rank}(d, y)}
 \pi(d \mid y_{1:k-1}) 
\mleft( \theta_k
\rho_d - \sum_{x=k}^K \theta_x
\rho_{y_x}
\mright)
\Bigg].
\end{split}
\label{eq:plrank2}
\end{equation}
The PL-Rank-2 algorithm~\citep{oosterhuis2021computationally} can approximate the gradient based on $N$ sampled rankings for $D$ items and a ranking length of $K$ with a computational complexity of:
$\mathcal{O}(NDK)$.
Table~\ref{table:properties} compares the theoretical properties of PL-Rank with other \ac{LTR} methods.
Importantly, PL-Rank is the only method that is not a pairwise method, while also being based on the actual metric that is optimized.\footnote{We define a pairwise method as any method that uses an iteration over item pairs in their algorithm. Thus under our definition methods like ApproxNDCG -- that approximates the ranks of items -- and SoftRank -- that approximates ranking via a distribution over ranks per item -- are considered pairwise methods because they iterate over all possible item pairs to compute their approximations and gradients.}
Furthermore, it shares most properties with StochasticRank: an alternative method for approximating the policy gradient.

\begin{algorithm}[t]
\caption{PL-Rank-3 Gradient Estimation} 
\label{alg:plrank3}
\begin{algorithmic}[1]
\STATE \textbf{Input}: items: $\mathcal{D}$; Relevances: $\rho$; Metric weights: $\theta$;
\\\phantom{ \textbf{Input}}
Score function: $f$; Number of samples: $N$. \label{alg:line:input}
\STATE $\{y^{(1)}, y^{(2)},\dots,y^{(N)}\} \leftarrow \text{Gumbel\_Sample}(N, m)$ \label{alg:line:gumbel}
\STATE $\text{Grad} \leftarrow \mathbf{0}$ \hfill \textit{\small // initialize zero gradient per item}  \label{alg:line:initlambda}
\STATE $S \leftarrow \sum_{d\in\mathcal{D}} e^{f(d)}$ \hfill \textit{\small // initialize PL denominator}  \label{alg:line:initdenom}
\FOR{$i \in [1,2,\ldots,N]$}
    \STATE $S' \leftarrow S$ \hfill \textit{\small // copy initial PL denominator} 
    \STATE $PR_{K+1} \leftarrow 0$ \hfill \textit{\small // zero value for easy computation} 
    \FOR{$k \in [K, K-1, \ldots,1]$}
    \STATE $PR_{k} \leftarrow PR_{k+1} + \theta_k\rho_{y_k^{(i)}}$ \label{alg:line:PR} \hfill \textit{\small // pre-compute all placement rewards}
    \ENDFOR
    \STATE $(RI_{0}, DR_{0}) \leftarrow (0, 0)$  \hfill \textit{\small // zero values for easy computation} 
    \FOR{$k \in [1, 2,\ldots, K]$}
    \STATE $RI_{k} \leftarrow RI_{k-1} + PR_k/S$  \label{alg:line:RI} \hfill \textit{\small // pre-compute all risks}
    \STATE $DR_{k} \leftarrow DR_{k-1} + \theta_k/S$  \label{alg:line:DR} \hfill \textit{\small // pre-compute all direct rewards}
    \STATE $S' \leftarrow S' - e^{f(y_k^{(i)})}$  \hfill \textit{\small // renormalize denominator} \label{alg:line:denomupdate}
    \ENDFOR
    \FOR{$d \in \mathcal{D}$}
        \STATE $r \leftarrow \min(\text{rank}(d,y), K)$ \hfill \textit{\small // index of the pre-computed values}
        \STATE $\text{Grad}(d) \leftarrow \text{Grad}(d) + \frac{1}{N}(PR_{r+1} + e^{f(d)}(\rho_dDR_r-RI_r))$ %
        \ENDFOR
\ENDFOR
\RETURN $\text{Grad}$ \label{alg:line:return}
\end{algorithmic}
\end{algorithm}

\setlength{\tabcolsep}{1pt}

{\renewcommand{\arraystretch}{0.5}
\begin{figure*}[tb]
\centering
\begin{tabular}{r r r}
 \multicolumn{1}{c}{ \hspace{0.4cm} Yahoo!\ Webscope-Set1}
&
 \multicolumn{1}{c}{ \hspace{0.1cm} MSLR-Web30k}
&
 \multicolumn{1}{c}{ \hspace{0.1cm} Istella}
\\
\rotatebox[origin=lt]{90}{\hspace{0.2cm} \small Minutes per Epoch} 
\includegraphics[scale=0.36]{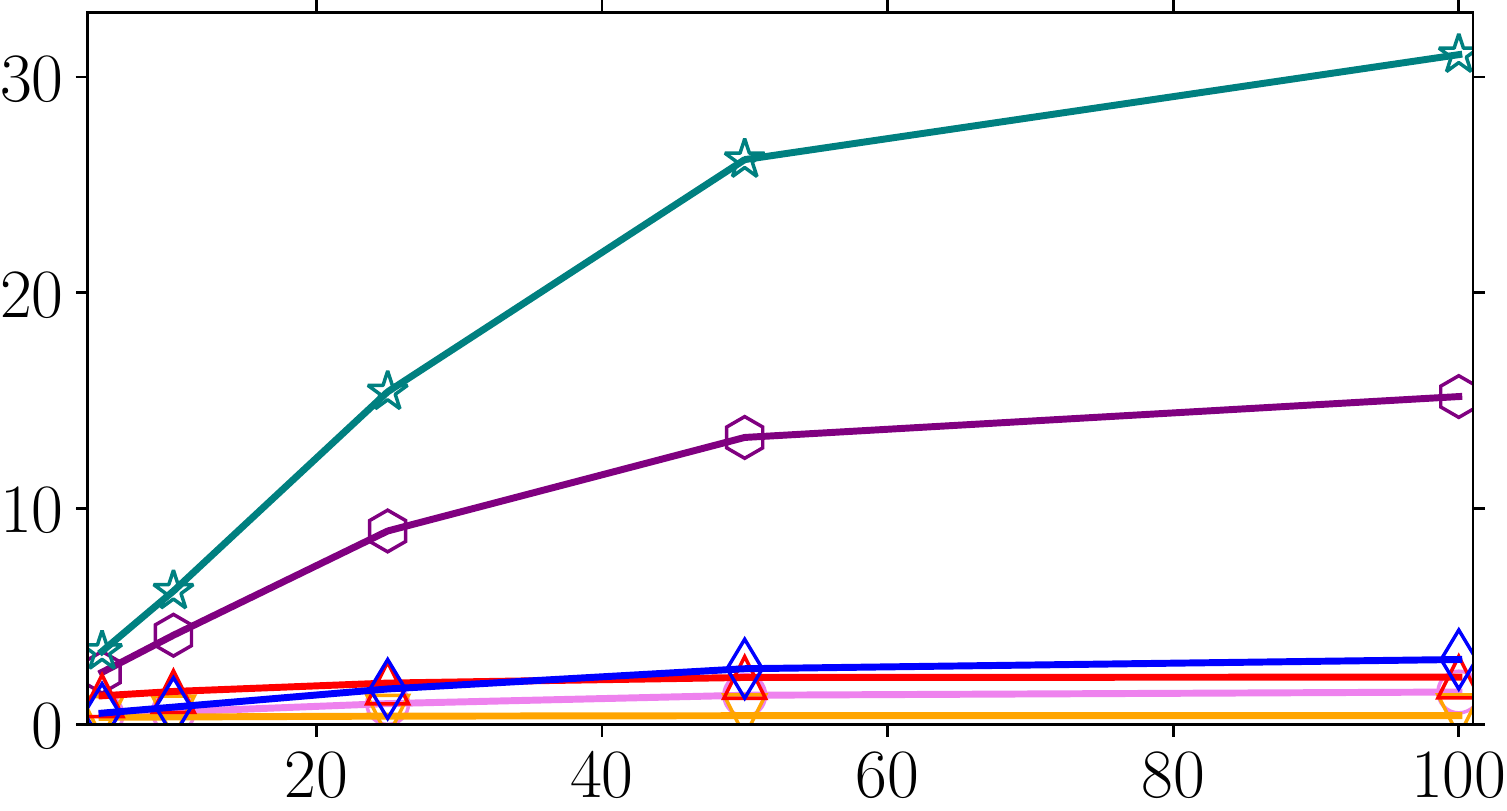}&
\includegraphics[scale=0.36]{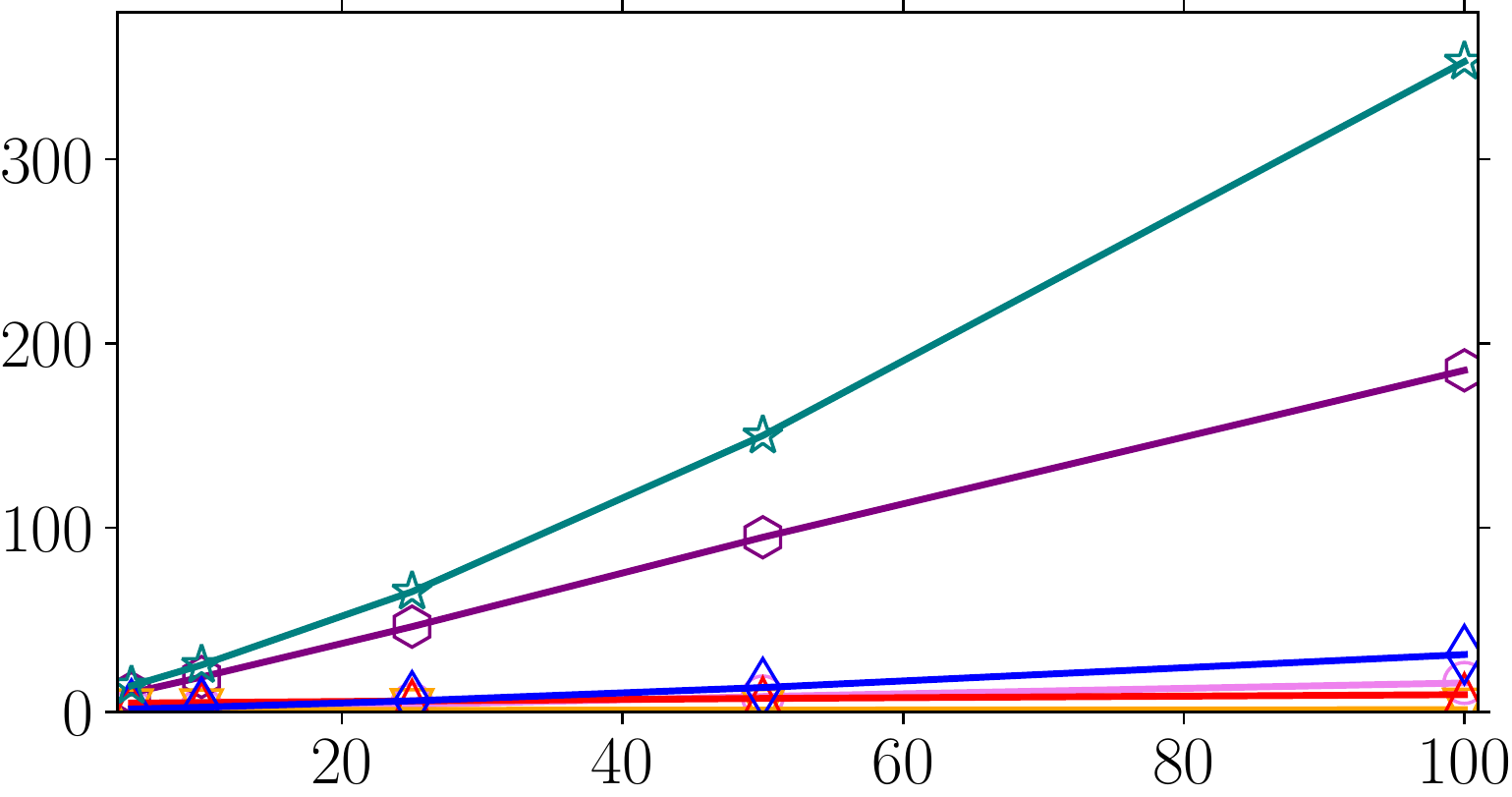} &
\includegraphics[scale=0.36]{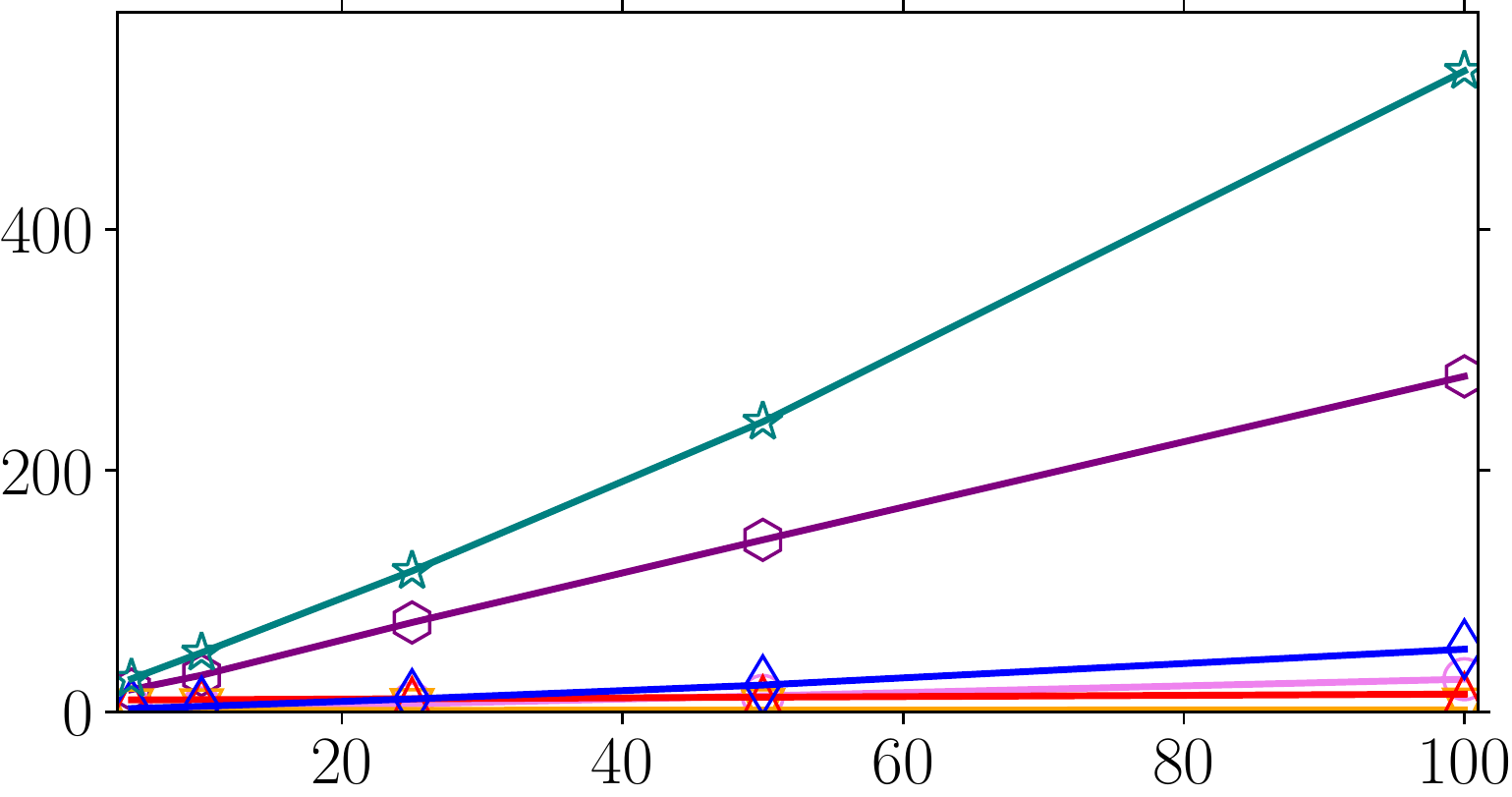}
\\
 \multicolumn{1}{c}{\small \hspace{0.5em} Cutoff $K$}
& \multicolumn{1}{c}{\small \hspace{0.5em} Cutoff $K$}
& \multicolumn{1}{c}{\small \hspace{0.5em} Cutoff $K$}
\\ 
\multicolumn{3}{c}{
\includegraphics[scale=.5]{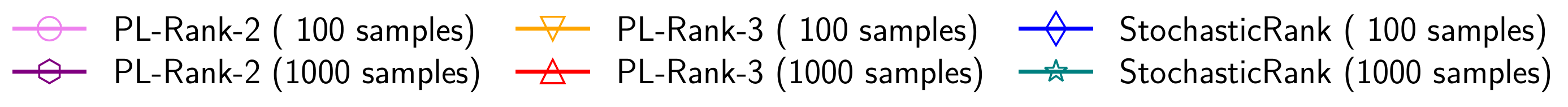}
} 
\end{tabular}
\caption{Mean number of minutes required to complete one training epoch as $K$ varies on three datasets.}
\vspace{0.4\baselineskip}
\label{fig:time}
\end{figure*}
}

{\renewcommand{\arraystretch}{0.5}
\begin{figure*}[tb]
\centering
\begin{tabular}{r r r}
 \multicolumn{1}{c}{  \hspace{0.95cm} Yahoo!\ Webscope-Set1}
&
 \multicolumn{1}{c}{  \hspace{0.4cm} MSLR-Web30k}
&
 \multicolumn{1}{c}{  \hspace{0.6cm} Istella}
\\
\rotatebox[origin=lt]{90}{\hspace{0.85cm} \small \emph{$K=5$}} 
\rotatebox[origin=lt]{90}{\hspace{0.6cm} \small NDCG@5} 
\includegraphics[scale=0.36]{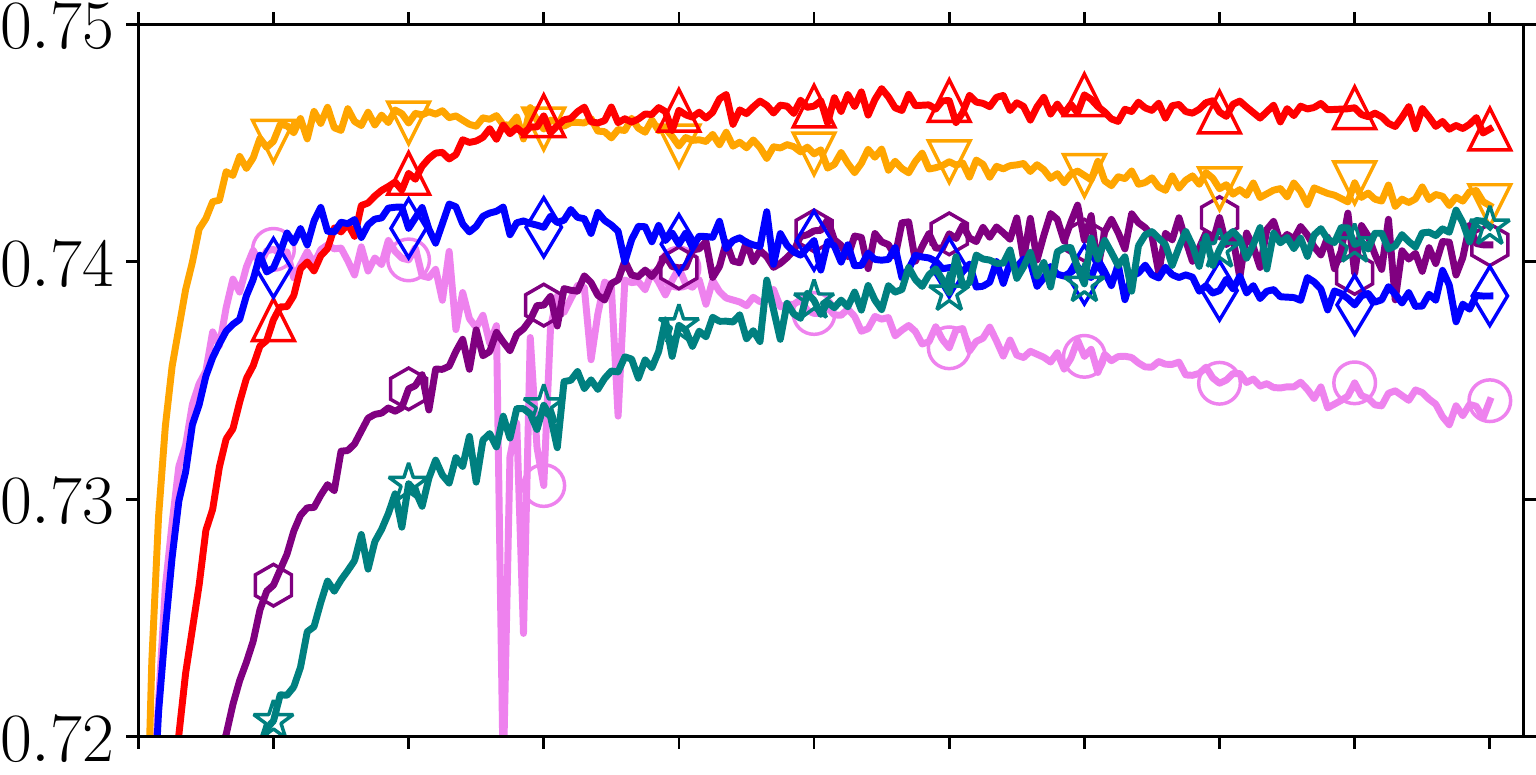}&
\includegraphics[scale=0.36]{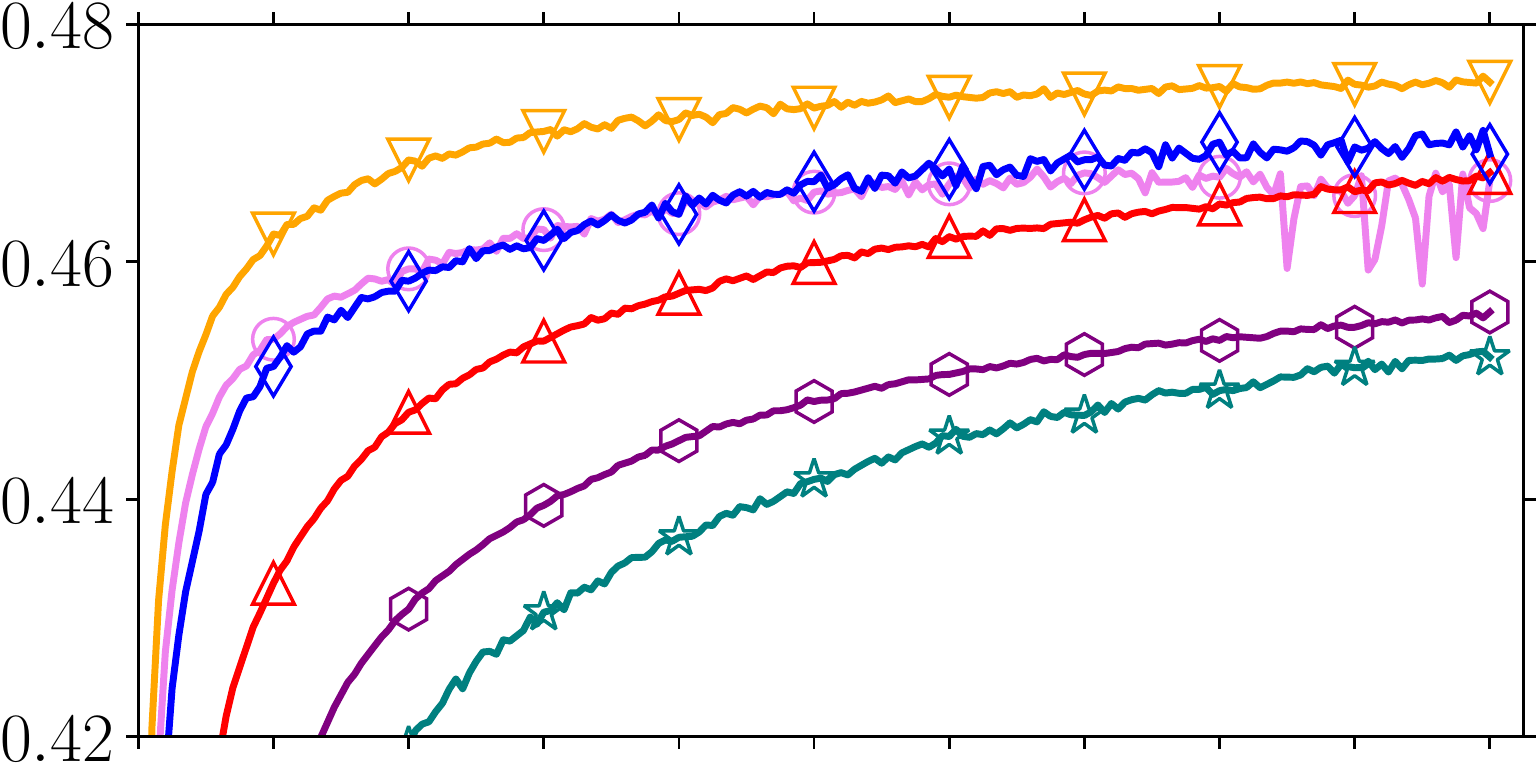} &
\includegraphics[scale=0.36]{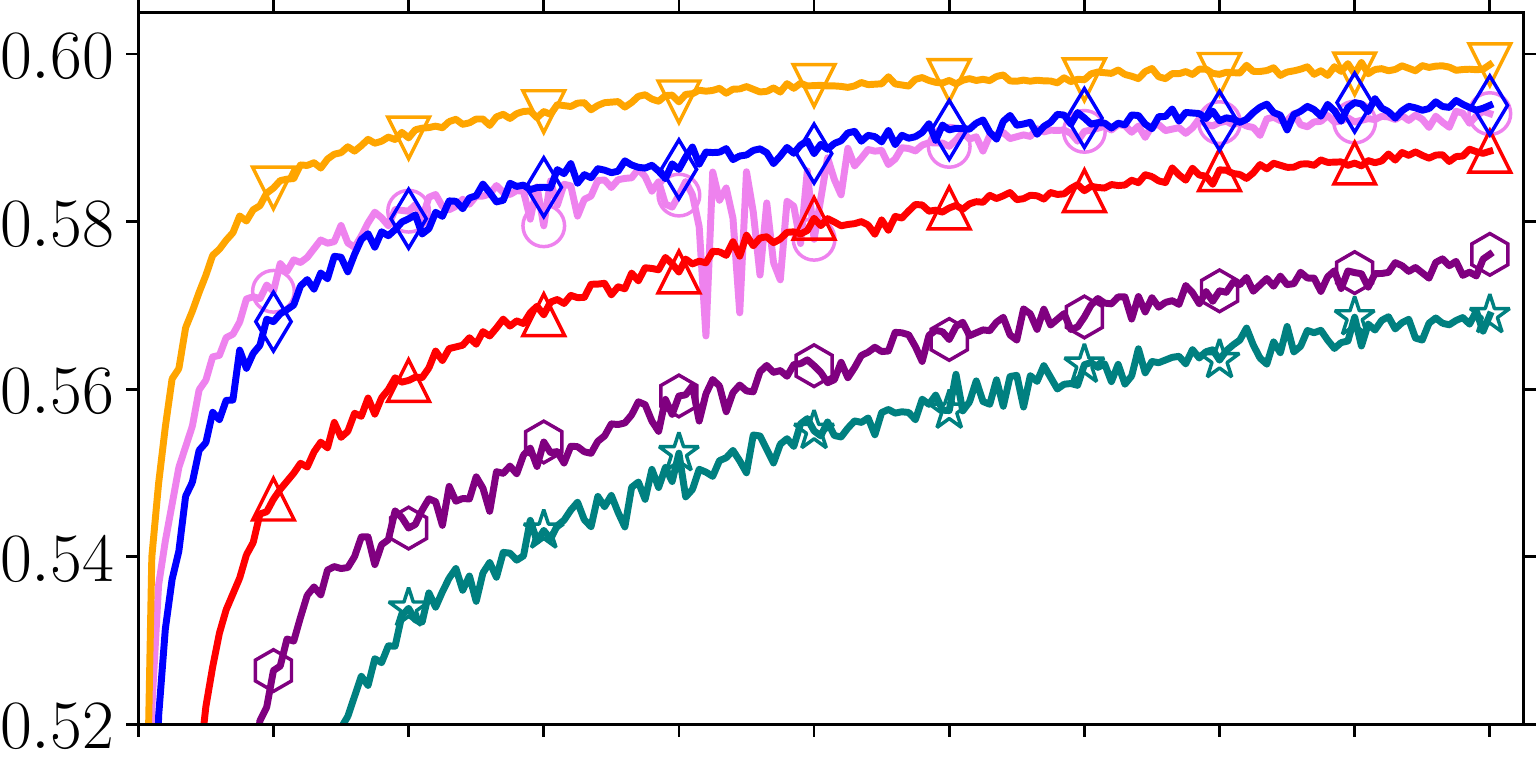}
\\
\rotatebox[origin=lt]{90}{\hspace{0.7cm} \small \emph{ $K=10$} }
\rotatebox[origin=lt]{90}{\hspace{0.52cm} \small NDCG@10} 
\includegraphics[scale=0.36]{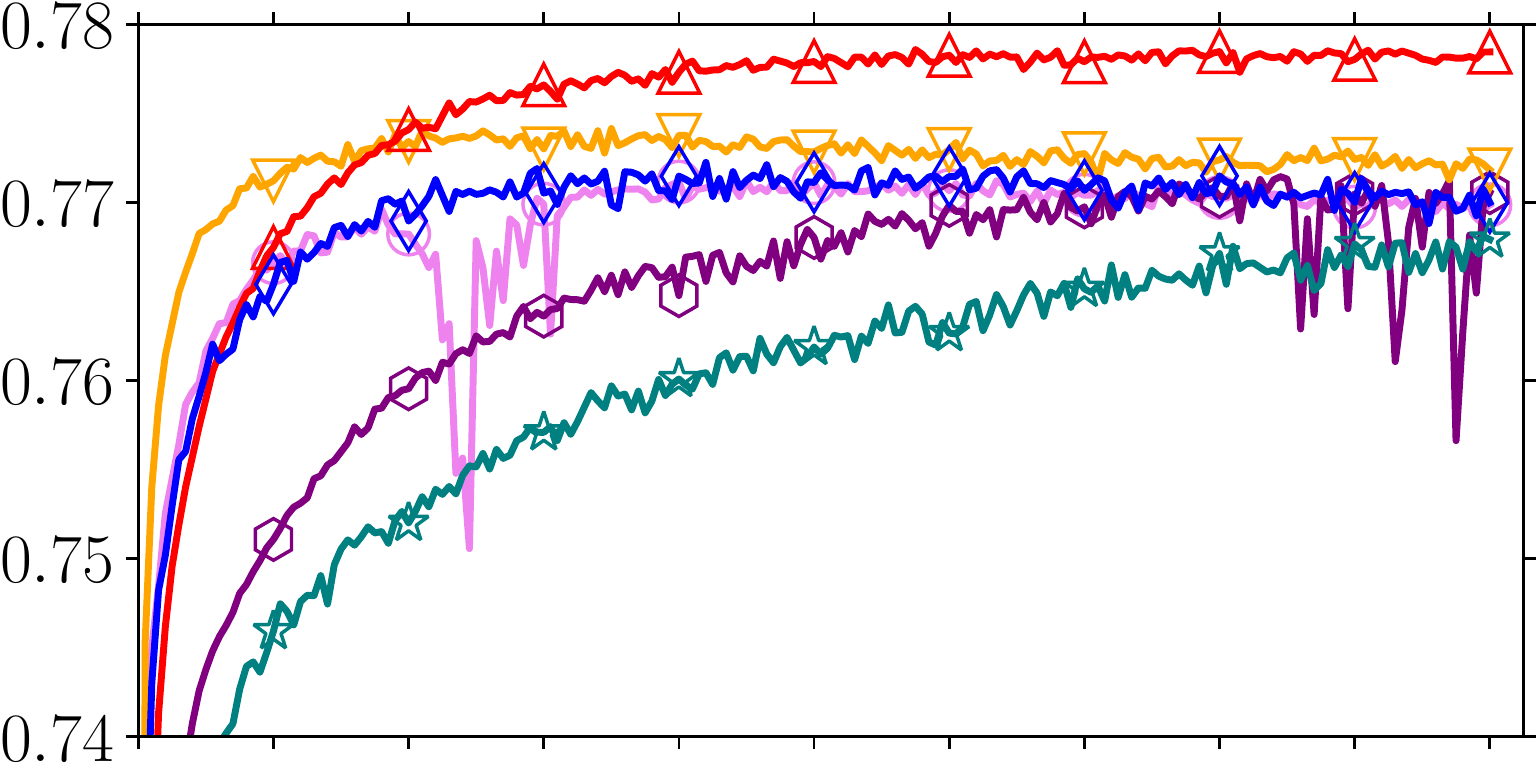} &
\includegraphics[scale=0.36]{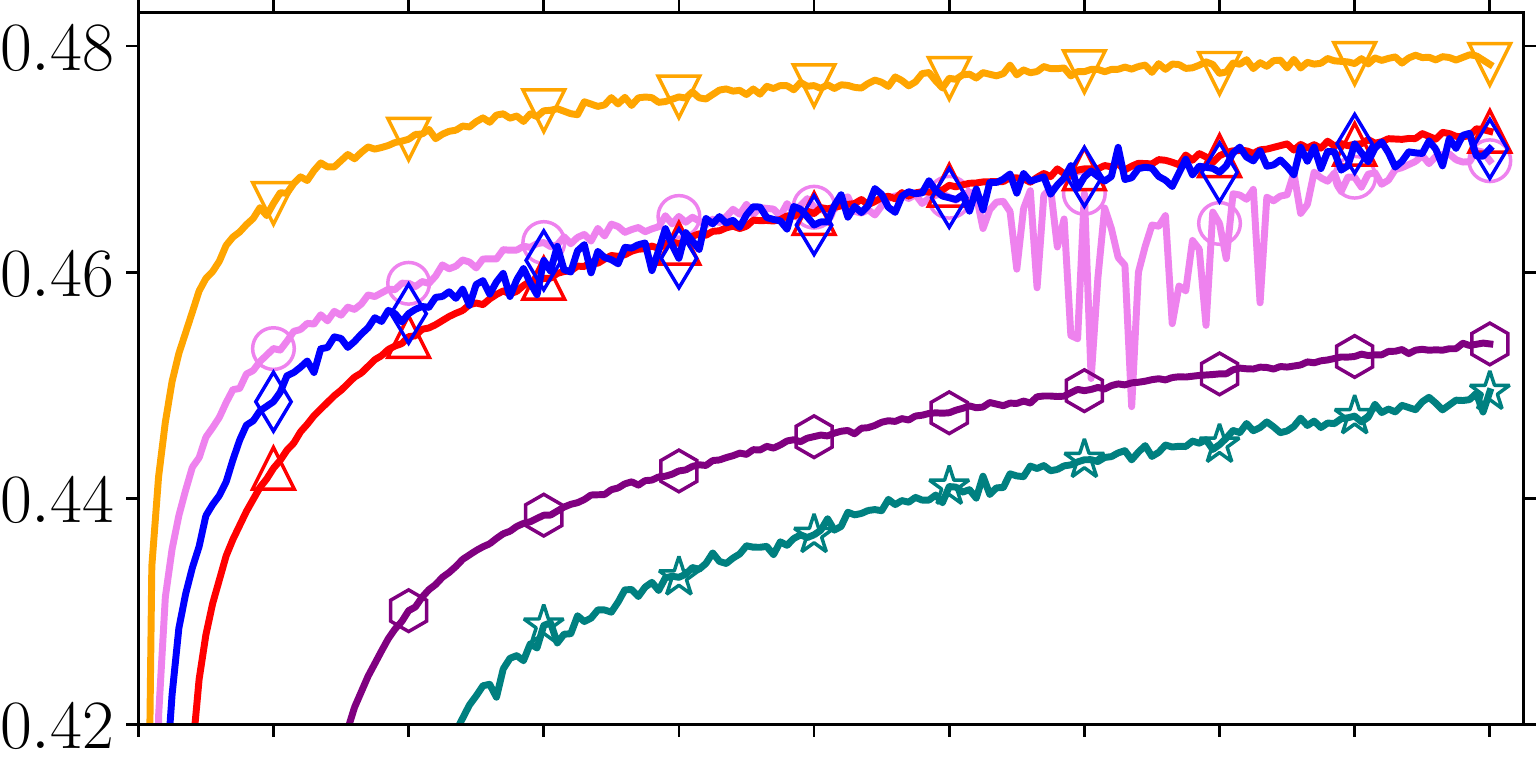} &
\includegraphics[scale=0.36]{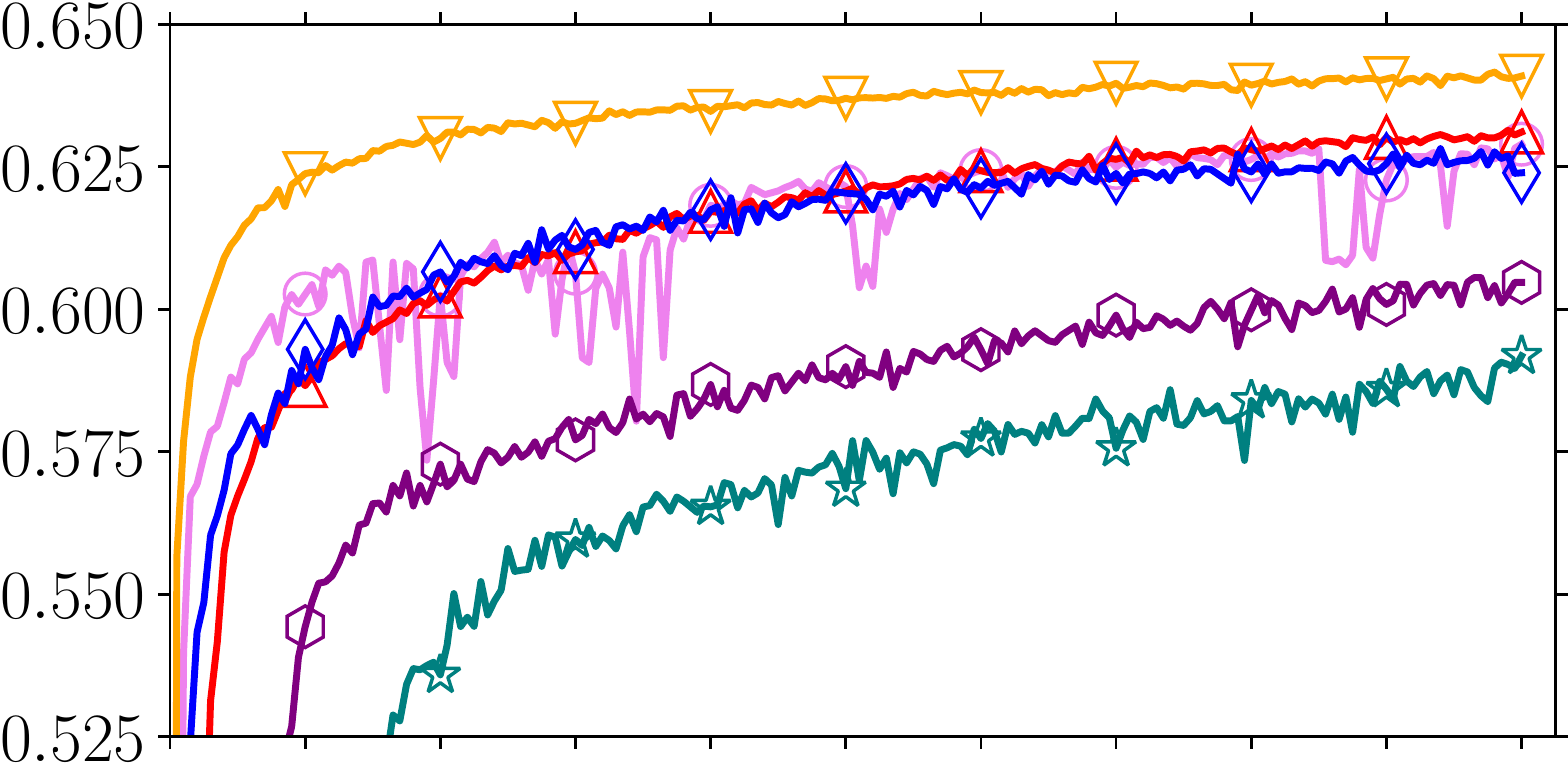}
\\
\rotatebox[origin=lt]{90}{\hspace{0.7cm} \small \emph{ $K=25$} }
\rotatebox[origin=lt]{90}{\hspace{0.52cm} \small NDCG@25} 
\includegraphics[scale=0.36]{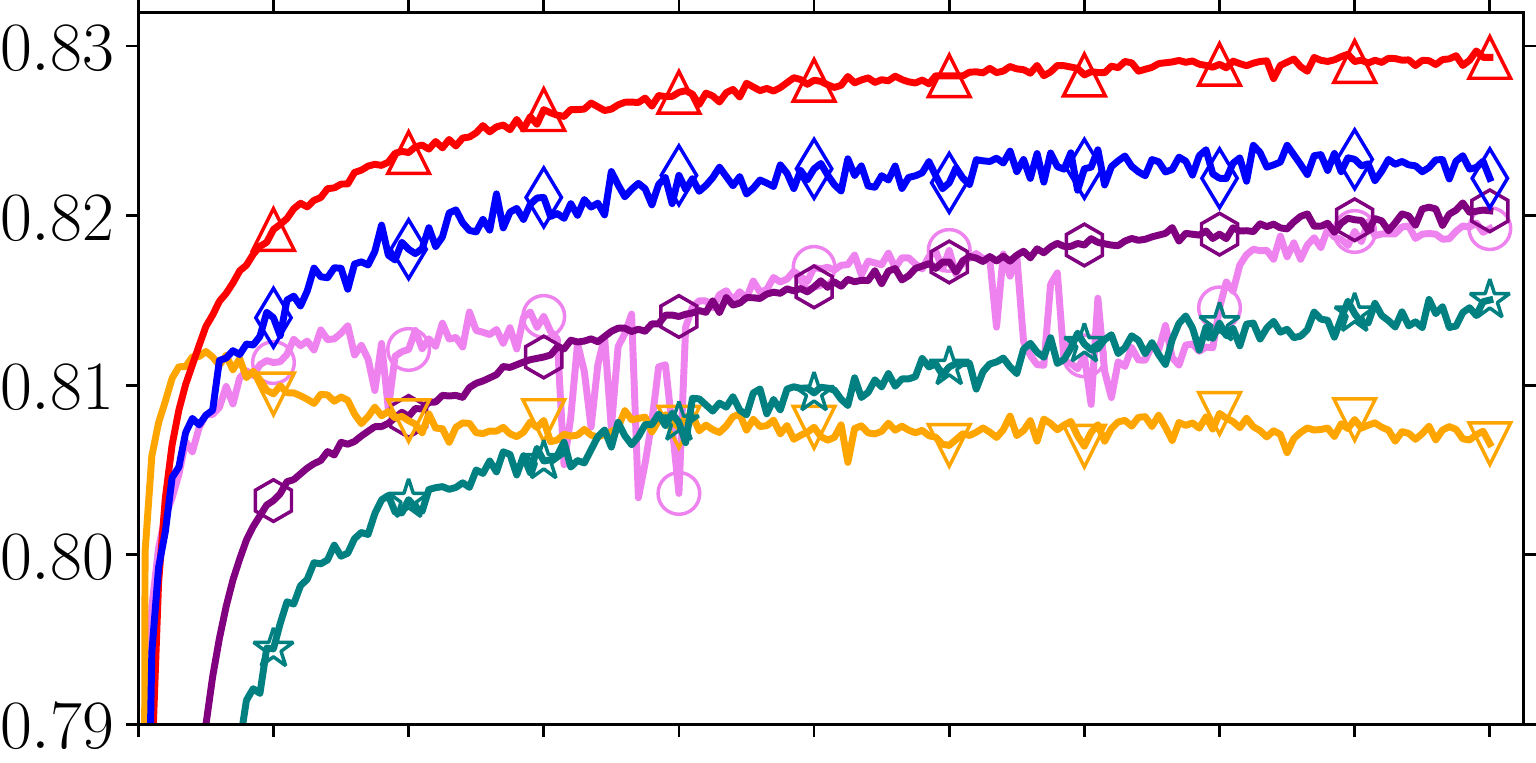} &
\includegraphics[scale=0.36]{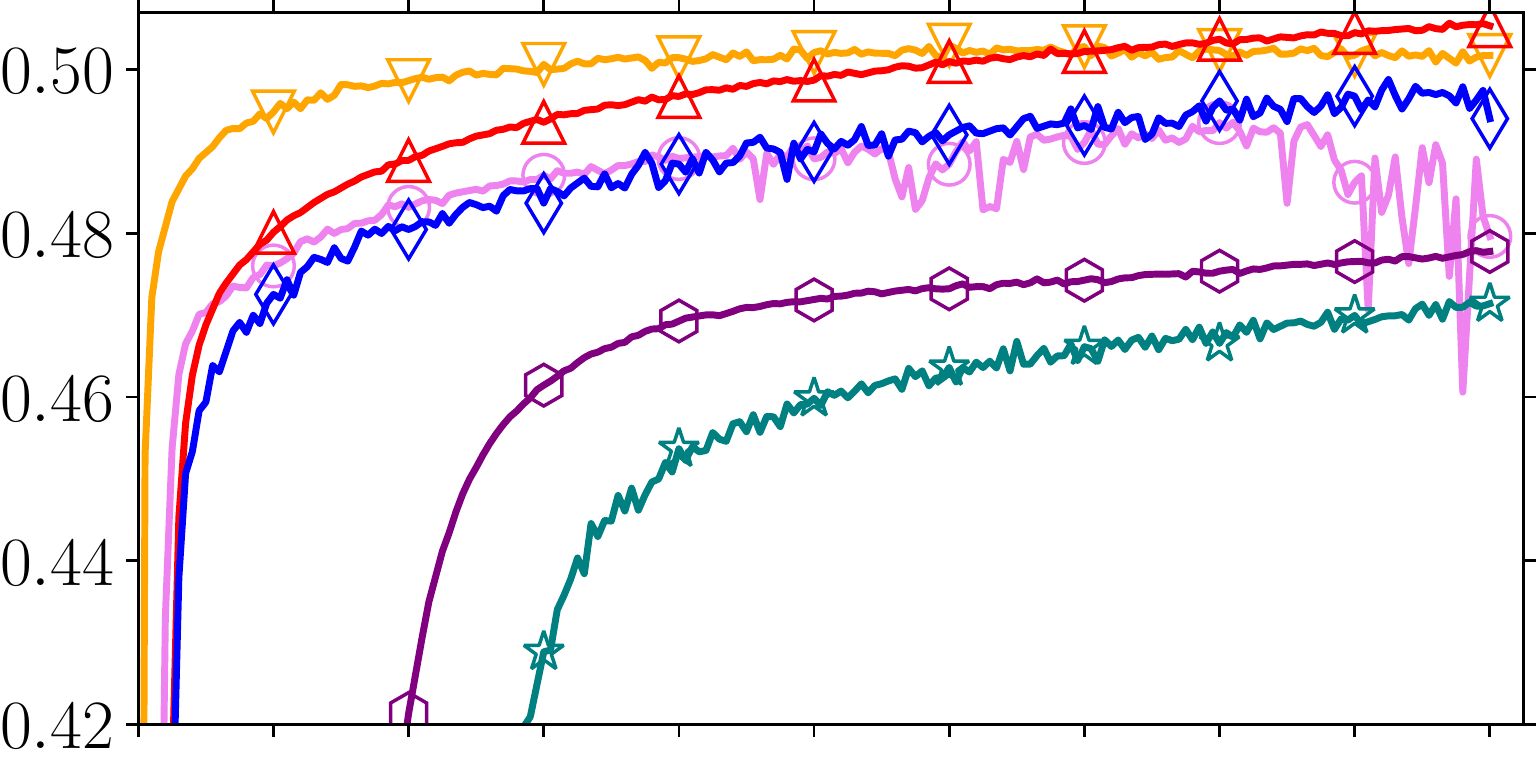} &
\includegraphics[scale=0.36]{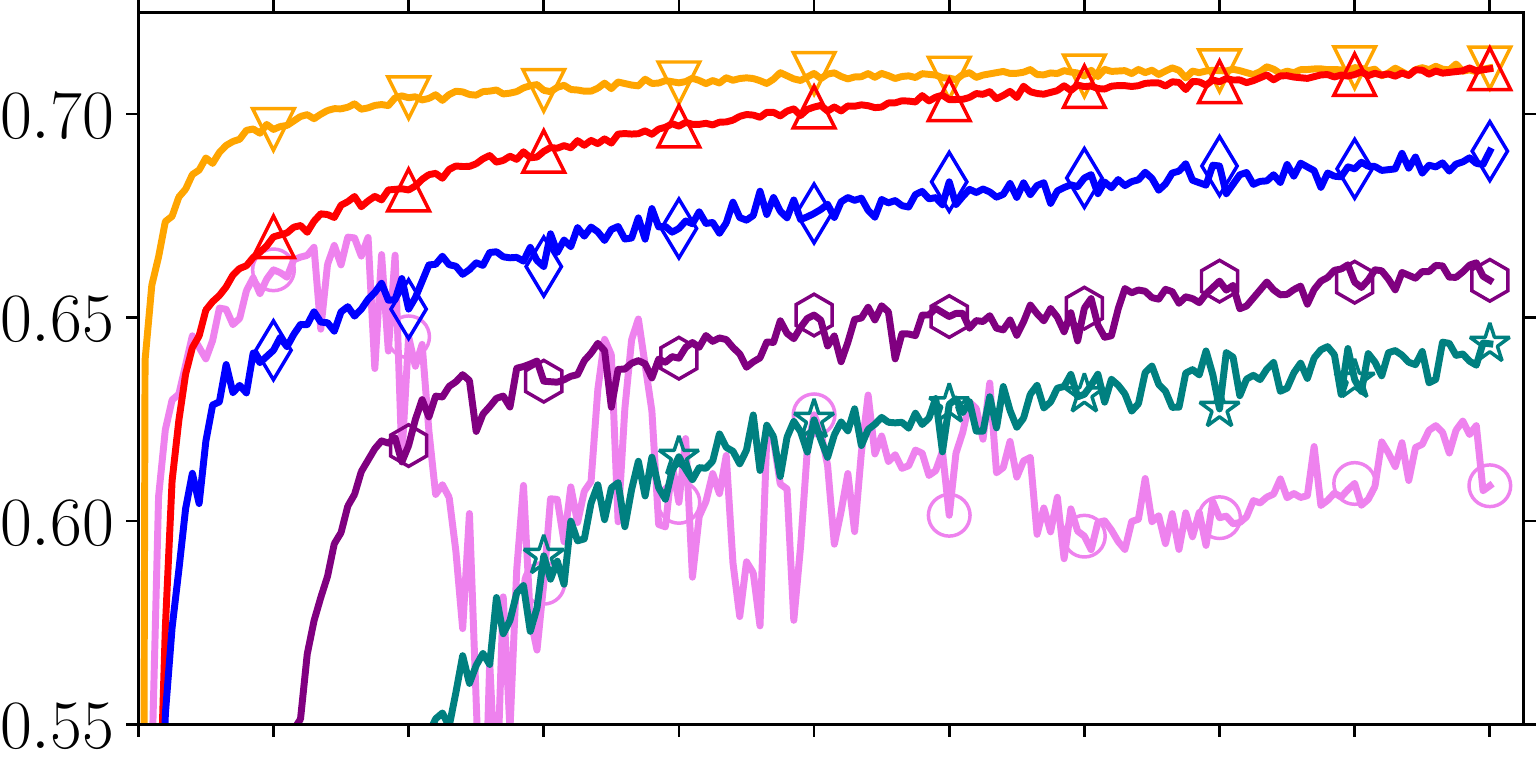}
\\
\rotatebox[origin=lt]{90}{\hspace{0.7cm}\small  \emph{ $K=50$}}
\rotatebox[origin=lt]{90}{\hspace{0.45cm} \small NDCG@50} 
\includegraphics[scale=0.36]{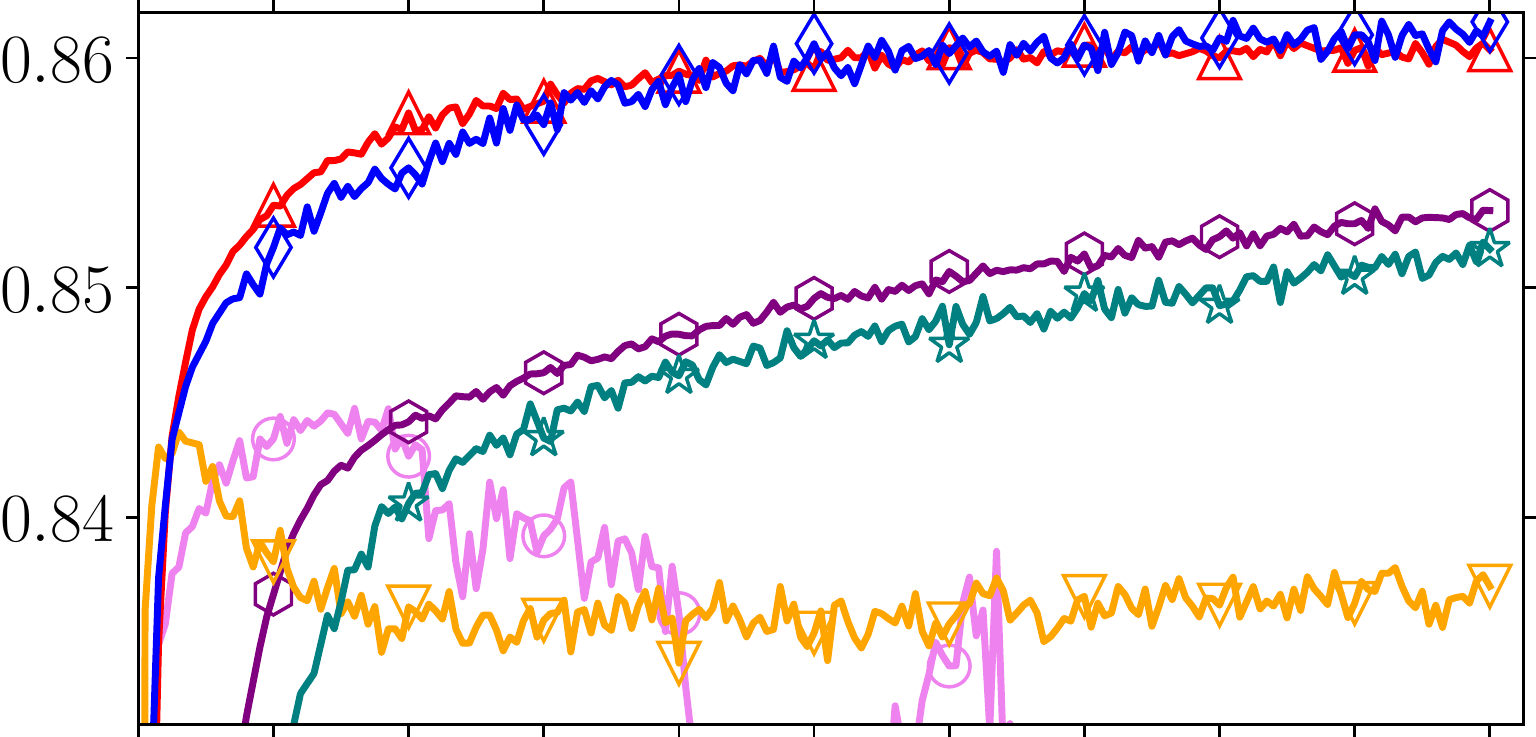} &
\includegraphics[scale=0.36]{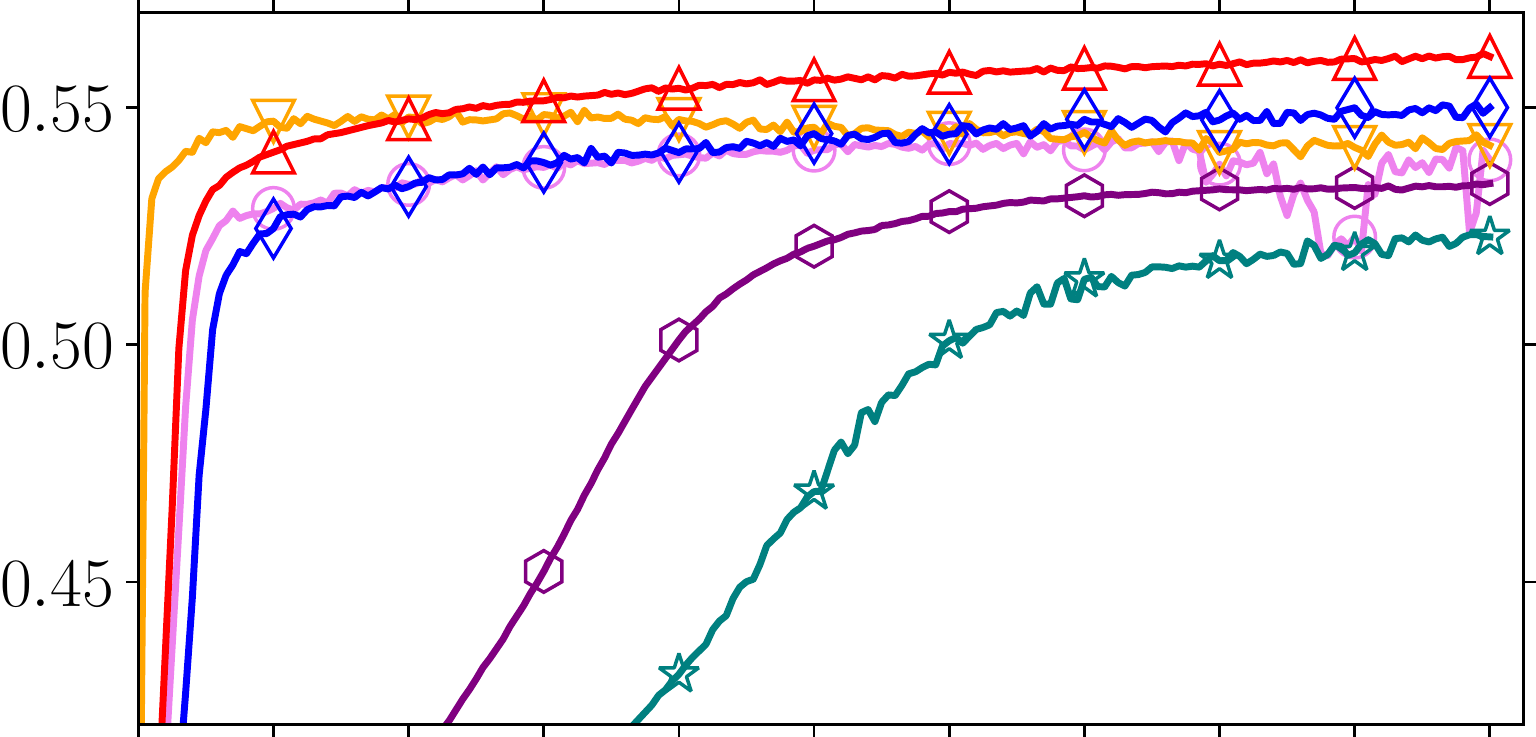} &
\includegraphics[scale=0.36]{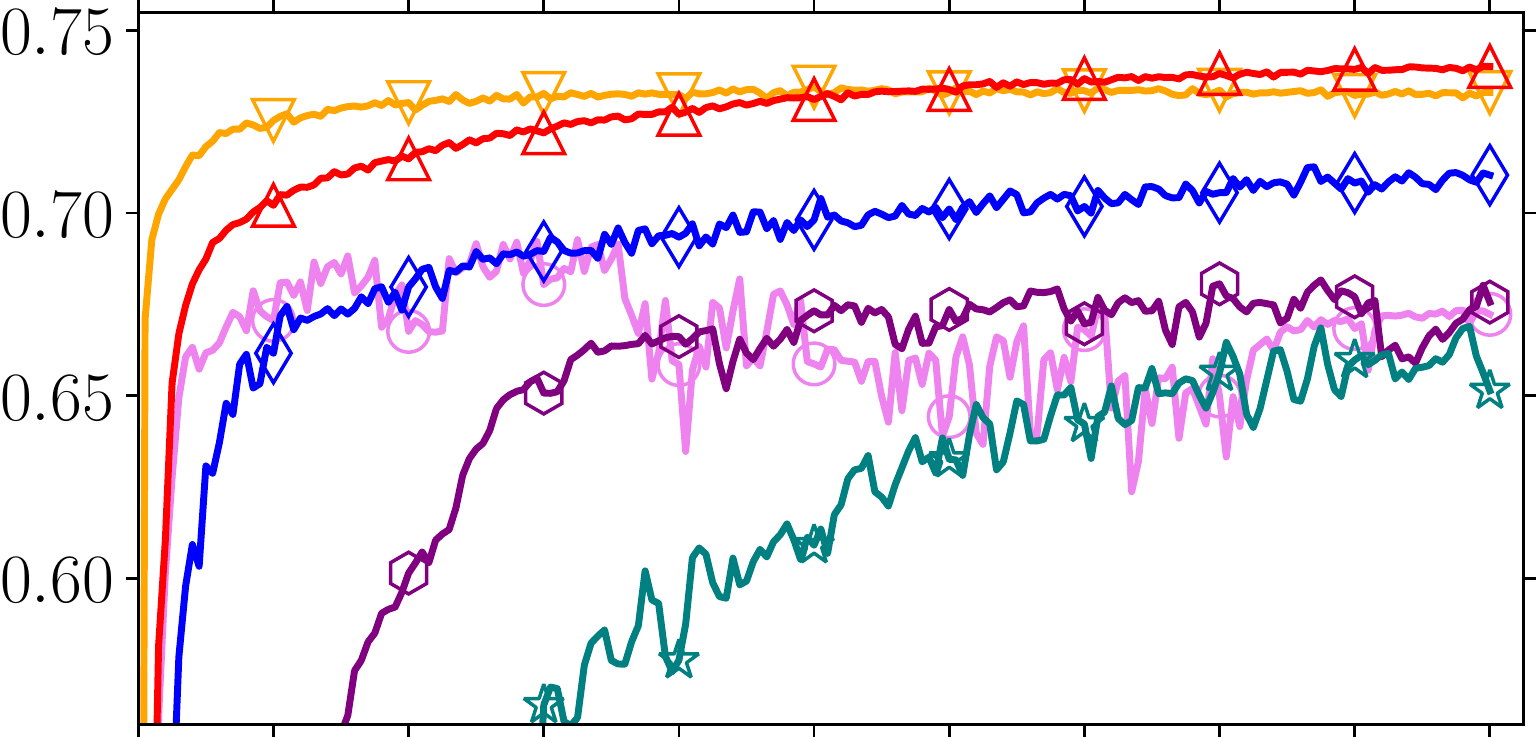}
\\
\rotatebox[origin=lt]{90}{\hspace{0.8cm} \small \emph{ $K=100$}}
\rotatebox[origin=lt]{90}{\hspace{0.58cm} \small NDCG@100} 
\includegraphics[scale=0.36]{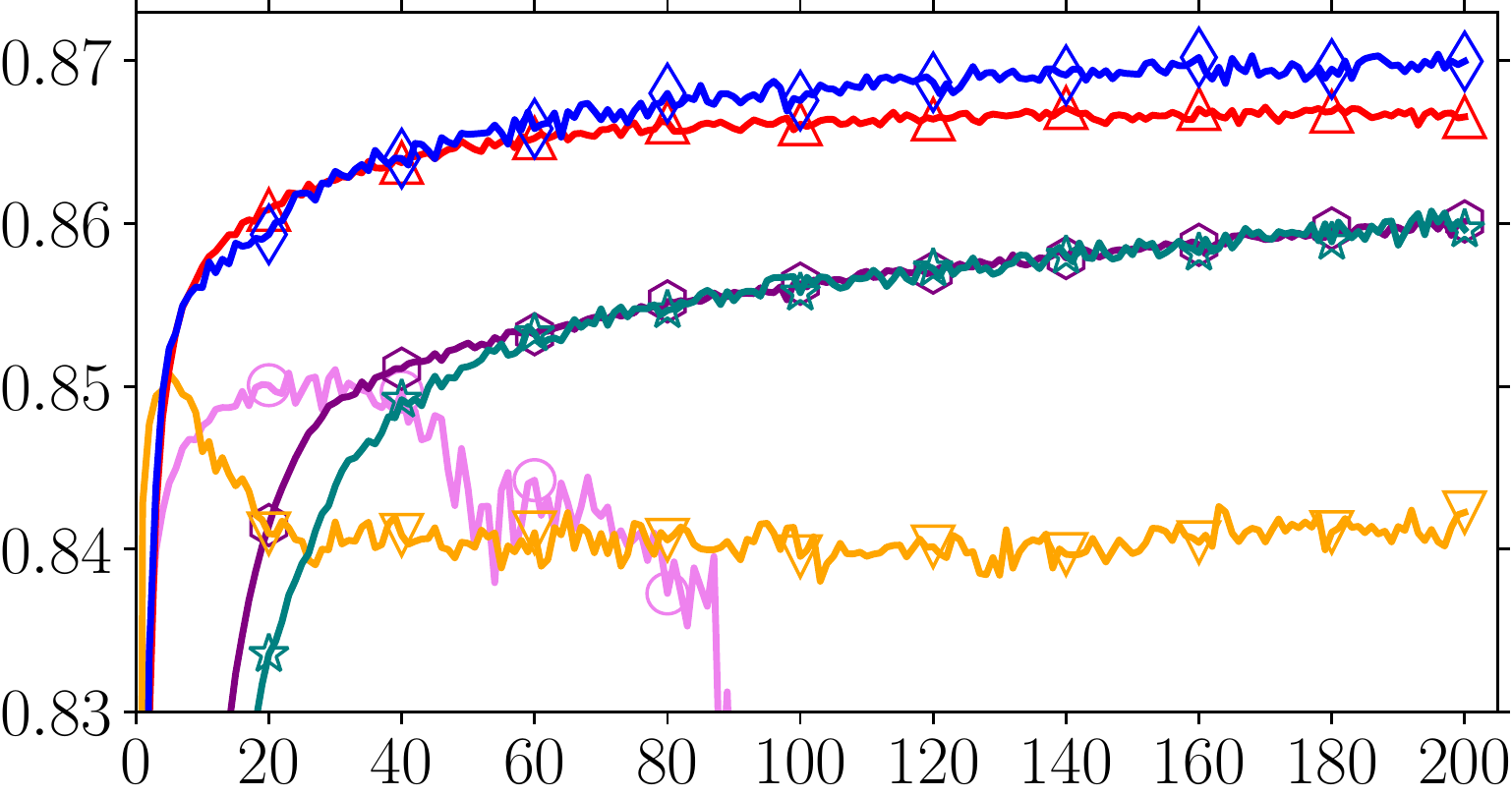} &
\includegraphics[scale=0.36]{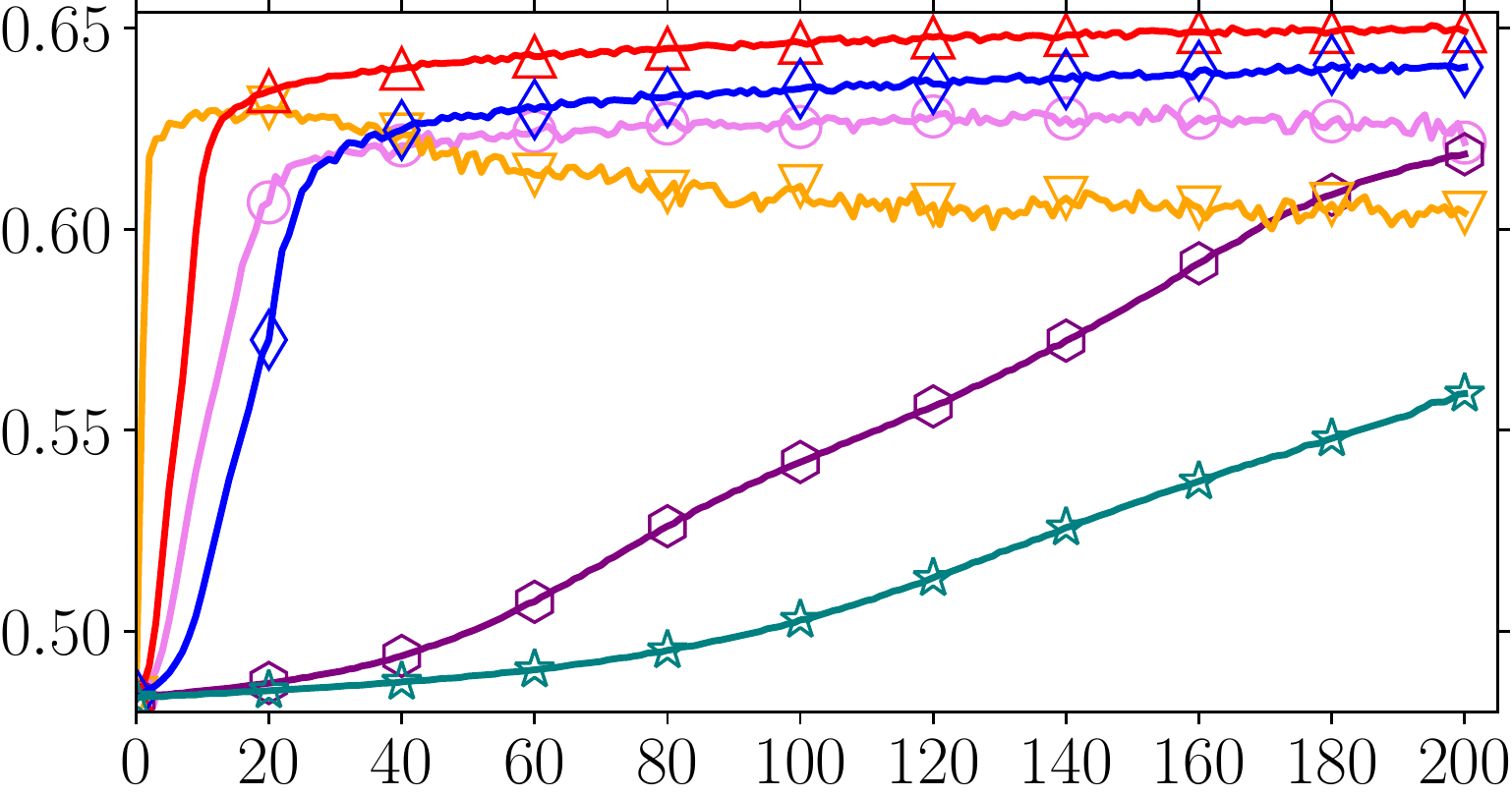} &
\includegraphics[scale=0.36]{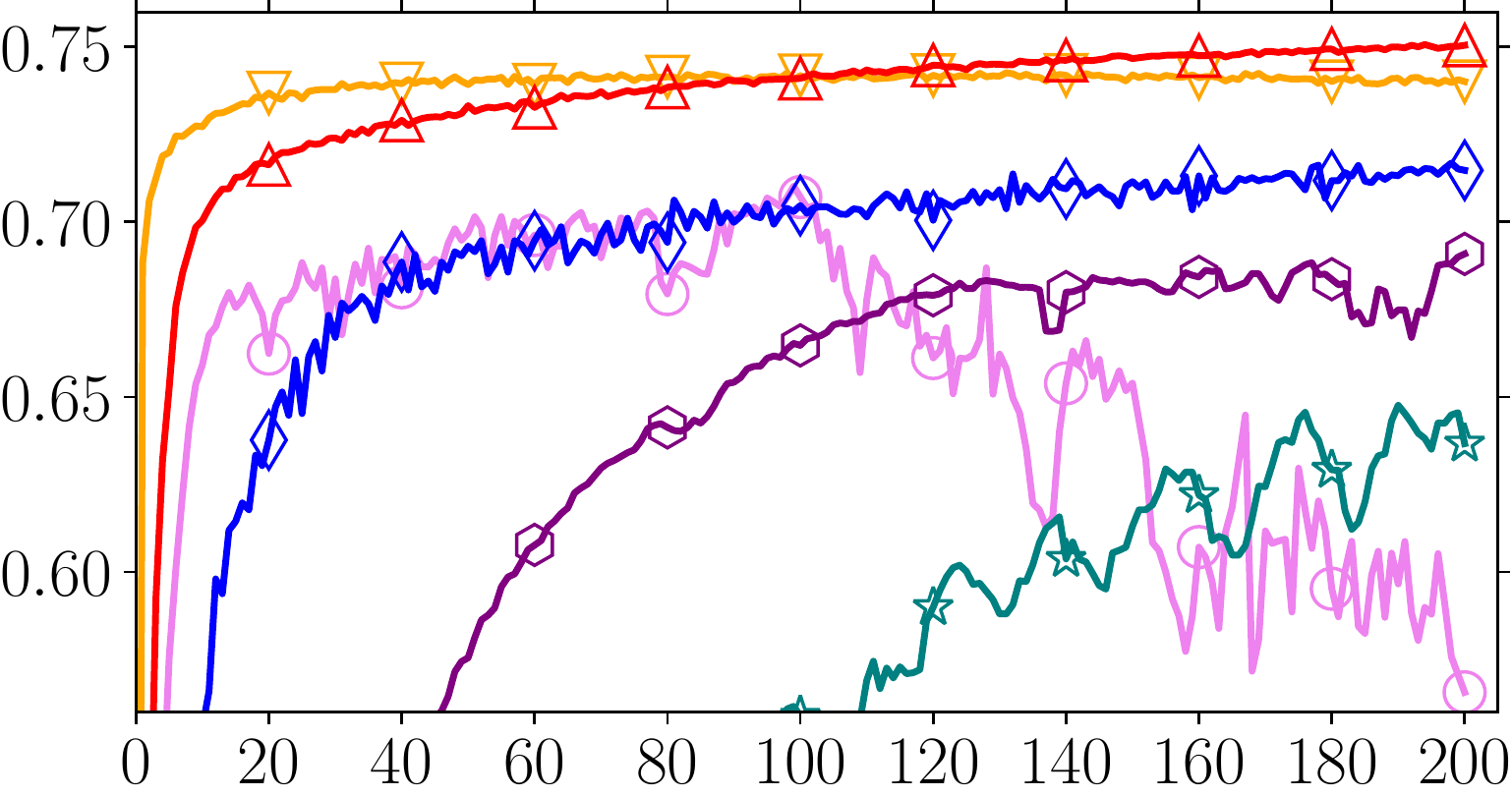}
\\
 \multicolumn{1}{c}{ \hspace{2.7em} Minutes Trained}
& \multicolumn{1}{c}{ \hspace{1.2em} Minutes Trained}
& \multicolumn{1}{c}{ \hspace{1.5em} Minutes Trained}
\\
\multicolumn{3}{c}{
\includegraphics[scale=.5]{figures/perf_legend}
} 
\end{tabular}
\caption{NDCG@K performance of PL-Rank-2, PL-Rank-3 and StochasticRank with $N=100$ and $N=1000$ when trained up to 200 minutes and for various $K$ values evaluated on the test-set of three datasets.
All displayed results are averages over 30 independent runs.
NDCG was normalized on dataset-level instead of query-level~\citep{ferrante2021towards}.
}
\label{fig:perf}
\end{figure*}
}

\begin{table}
\caption{Mean number of minutes taken to complete one epoch with various methods, ranking length/cutoff $K$ and number of samples $N$ on three datasets.
Bold numbers indicate the minimal time per $N$ and $K$ values on each dataset.
}
\label{tab:timing}
\begin{tabularx}{\columnwidth}{c X r r r r r r}
\hline
\hspace{0.4cm}
& \multicolumn{1}{ c }{ method}
& \multicolumn{1}{ c }{$N$}
& \multicolumn{1}{ c }{\;\;$K=5$}
& \multicolumn{1}{ c }{ \;\;$K=10$}
& \multicolumn{1}{ c }{ \;\;$K=25$}
& \multicolumn{1}{ c }{ \;\;$K=50$}
& \multicolumn{1}{ c }{ \;\;$K=100$}
\\
\hline
\multirow{6}{*}{\rotatebox[origin=lt]{90}{\small Yahoo!}}
& \multirow{2}{*}{ Stoc.Rank} & \phantom{1}100
& 0.52
& 0.80
& 1.64
& 2.58
& 3.01
\\
&  & 1000
& 3.38
& 6.18
& 15.41
& 26.17
& 31.03
\\ \cline{3-8}
& \multirow{2}{*}{ PL-Rank-2} & \phantom{1}100
& 0.43
& 0.58
& 0.96
& 1.35
& 1.50
\\
& & 1000 
& 2.41
& 4.13
& 8.96
& 13.30
& 15.19
\\\cline{3-8}
& \multirow{2}{*}{ PL-Rank-3} & \phantom{1}100
& \bf 0.33
& \bf 0.34
& \bf 0.38
& \bf 0.40
& \bf 0.40
\\
& & 1000
& \bf 1.33
& \bf 1.53
& \bf 1.92
& \bf 2.18
& \bf 2.19
\\
\hline
\multirow{6}{*}{\rotatebox[origin=lt]{90}{\small MSLR}}
& \multirow{2}{*}{Stoc.Rank} & \phantom{1}100
& 1.55
& 2.53
& 6.01
& 13.23
& 31.12
\\
&  & 1000
& 14.15
& 25.46
& 65.25
& 150.00
& 353.01
\\ \cline{3-8}
& \multirow{2}{*}{PL-Rank-2} & \phantom{1}100
& 1.19
& 1.75
& 4.04
& 8.36
& 15.71
\\
& & 1000 
& 10.34
& 18.91
& 46.27
& 94.81
& 185.42
\\ \cline{3-8}
& \multirow{2}{*}{PL-Rank-3} & \phantom{1}100
& \bf 0.74
& \bf 0.77
& \bf 0.86
& \bf 1.00
& \bf 1.20
\\
& & 1000
& \bf 4.77
& \bf 5.09
& \bf 5.93
& \bf 7.31
& \bf 9.37
\\ 
\hline
\multirow{6}{*}{\rotatebox[origin=lt]{90}{\small Istella}}
& \multirow{2}{*}{Stoc.Rank} & \phantom{1}100
& 2.79
& 4.61
& 10.75
& 22.25
& 51.94
\\
&  & 1000
& 27.34
& 49.07
& 116.73
& 240.40
& 531.09
\\ \cline{3-8}
& \multirow{2}{*}{PL-Rank-2} & \phantom{1}100
& 1.96
& 2.87
& 6.72
& 13.25
& 27.07
\\
& & 1000 
& 18.49
& 30.37
& 74.09
& 142.55
& 278.01
\\ \cline{3-8}
& \multirow{2}{*}{PL-Rank-3} & \phantom{1}100
& \bf 1.24
& \bf 1.26
& \bf 1.34
& \bf 1.46
& \bf 1.72
\\
& & 1000
& \bf 9.93
& \bf 10.14
& \bf 10.87
& \bf 12.20
& \bf 14.75
\\
\hline
\end{tabularx}
\end{table}

\section{Method: A Faster PL-Rank Algorithm}

We will now introduce PL-Rank-3: an algorithm for computing the same approximation as PL-Rank-2 but with a significantly better computational complexity.
Algorithm~\ref{alg:plrank3} displays PL-Rank-3 in pseudo-code, we will now show that it produces the same approximation as PL-Rank-2 (cf.\ Algorithm~1 in \citep{oosterhuis2021computationally}).

To start, we define $PR_{y,i}$ as the placement reward, the reward following the item placed at rank $i$ in ranking $y$:
\begin{equation}
PR_{y,i} = \sum_{k=i}^{\min(i,K)}
\theta_k
\rho_{y_k},  
\quad
PR_{y,d} = PR_{y,rank(d, y)+1}. 
\end{equation}
Importantly, all $PR_{y,i}$ values for a ranking can be computed in $K$ steps (Line~\ref{alg:line:PR}).
Next, we point out that a summation over item placement probabilities similar to Eq.~\ref{eq:plrank2} can be formulated as:
\begin{equation}
\sum_{k=1}^{\text{rank}(d, y)}\hspace{-0.1cm}
 \pi(d \mid y_{1:k-1})
 =
 e^{f(d)} \mleft(\sum_{k=1}^{\text{rank}(d, y)} \frac{1}{\sum_{d' \in D \setminus y_{1:k-1}} e^{f(d')}}  \mright).
\end{equation}
Using this insight, we define $RI_{y,i}$ to later compute the \emph{risk} imposed by items with a non-zero placement probability at rank $i$ in $y$:
\begin{equation}
RI_{y,i} = \sum_{k=1}^{\min(i,K)} \frac{PR_{y,k}}{\sum_{d' \in D \setminus y_{1:k-1}} e^{f(d')}},
\quad RI_{y,d} = RI_{y,\text{rank}(d, y)}.
\end{equation}
Similarly, we define $DR_{y,i}$ to compute the expected direct reward:
\begin{equation}
DR_{y,i} = \sum_{k=1}^{\min(i,K)} \frac{\theta_k}{\sum_{d' \in D \setminus y_{1:k-1}} e^{f(d')}},
\; DR_{y,d} = DR_{y,\text{rank}(d, y)}.
\end{equation}
Crucial is that all $RI_{y,i}$ and $DR_{y,i}$ values can also be computed in $K$ steps (Line~\ref{alg:line:RI} \&~\ref{alg:line:DR}).
With these newly defined variables, we can reformulate Eq.~\ref{eq:plrank2} without any summation over $K$:
\begin{equation}
\frac{\delta \mathcal{R}(q)}{\delta f(d)} 
=
\mathbb{E}_{y} \mleft[
PR_{y,d} + e^{f(d)}\big( \rho_d DR_{y,d} - RI_{y,d}\big)
\mright].
\end{equation}
Accordingly, Algorithm~\ref{alg:plrank3} approximates the gradient using:
\begin{equation}
\begin{split}
\frac{\delta \mathcal{R}(q)}{\delta f(d)} 
\approx&
\frac{1}{N}\sum_{i=1}^N
\mleft(
PR_{y^{(i)},d} + e^{f(d)}\mleft( \rho_d DR_{y^{(i)},d} - RI_{y^{(i)},d}\mright)
\mright).
\end{split}
\end{equation}
PL-Rank-3 computes the $PR$, $RI$ and $DR$ values in $K$ steps and then reuses them for each of the $D$ items, resulting in a computational complexity of
$\mathcal{O}(N(D + K))$
given $N$ sampled rankings.
However, when we consider that sampling a ranking relies on (partial) sorting, the full complexity of applying PL-Rank-3 becomes
$\mathcal{O}(N(D + K\log(D))$.

Table~\ref{table:properties} reveals that - to the best of our knowledge - PL-Rank-3 has the best computational complexity of all metric-based \ac{LTR} methods.
Moreover, because its computational complexity is limited by the underlying sorting procedure, we posit that PL-Rank-3 has reached the minimum order of computational complexity that is possible for a \ac{LTR} method that is based on the full-ranking behavior of the model it optimizes.

\section{Experimental Setup}

We experimentally evaluate how the improvements in computational complexity translate to improvements in practical costs.
Our experimental runs optimize the $DCG@K$ of neural ranking models on the Yahoo!\ Webscope-Set1~\citep{Chapelle2011}, MSLR-Web30k~\citep{qin2013introducing} and Istella~\citep{dato2016fast} datasets.
The neural models have two-hidden layers of 32 sigmoid activation nodes, backpropagation via standard gradient descent with a learning rate of $0.01$ was applied using \emph{Tensorflow}~\citep{abadi2016tensorflow}.
We compare our PL-Rank-3 algorithm, with PL-Rank-2~\citep{oosterhuis2021computationally} and StochasticRank~\citep{ustimenko2020stochasticrank}.
StochasticRank was chosen as a baseline because it shares many properties with PL-Rank (see Table~\ref{table:properties}) yet was not compared with PL-Rank in previous work~\citep{oosterhuis2021computationally}.
Our StochasticRank implementation uses the Gumbel distribution as stochastic noise instead of the normal distribution of the original algorithm~\citep{ustimenko2020stochasticrank}, this makes the method applicable and effective to optimize a \ac{PL} model.
We reimplemented the PL-Rank and StochasticRank algorithms in \emph{Numpy}~\citep{harris2020array} and performed all our experiments on
AMD EPYC™ 7H12 CPUs.\footnote{https://www.amd.com/en/products/cpu/amd-epyc-7H12}
The ranking length/metric cutoff was varied: $K\in\{5,10,25,50,100\}$ and number of sampled rankings: $N\in\{100, 1000\}$, to measure their impact on computational costs.
Performance was measured over 200 minutes of training time, in addition to the average time required to complete one training epoch.
All reported results are averages over 30 independent runs performed under identical circumstances.
We report Normalized $DCG@K$ (NDCG@K) as our ranking performance metric computed on the held-out test-set of each dataset; following the advice of~\citet{ferrante2021towards}, we do not use query-normalization but dataset-normalization: we divide the $DCG@K$ of a ranking model by the maximum possible $DCG@K$ value on the entire test-set of the dataset.

\section{Results}

Figure~\ref{fig:time} and Table~\ref{tab:timing} shows the effect of ranking length $K$ on the computational costs of the \ac{LTR} methods.
As expected, Figure~\ref{fig:time} reveals that PL-Rank-2 and StochasticRank are heavily affected by increases in $K$: there appears to be a clear linear trend on the MSLR and Istella datasets.
We note that many queries in the Yahoo!\ dataset have less than 100 documents ($D\leq100$) which could explain why the effect is sub-linear on that dataset.
In contrast, PL-Rank-3 appears barely affected by $K$ on all of the datasets.

Table~\ref{tab:timing} allows us to also compare the computational costs in more detail.
Regardless of whether $N=100$ or $N=1000$, the required times of PL-Rank-2 and StochasticRank scale close to linearly with $K$, but those for PL-Rank-3 do not.
For instance, on Istella with $N=100$ and $K=5$, PL-Rank-2 needs 1.96 minutes, StochasticRank needs 2.79 and PL-Rank-3 needs 1.24 minutes, when compared to $K=100$, PL-Rank-2 needs an additional 25 minutes, StochasticRank 49 minutes more but PL-Rank-3 only requires an increase of 28 seconds. 
Moreover, PL-Rank-3 has the lowest computational costs when compared to the other methods with the same $N$ and $K$ values, across all three datasets.
We thus conclude that in terms of time required to complete a single epoch, PL-Rank-3 is a clear improvement over PL-Rank-2 and StochasticRank.
Additionally, in terms of scalability of computational costs w.r.t.\ $K$, PL-Rank-3 is the best choice by a considerable margin.

Now that we have established that PL-Rank-3 completes training epochs considerably faster than the other methods, we consider its effect on how quickly certain performance can be reached.
Figure~\ref{fig:perf} show the performance of the \ac{LTR} methods when optimizing $DCG@K$ over 200 minutes for various $K$ values.
In all but two scenarios, PL-Rank-3 provides the highest performance at all times, where it seems to mostly depend on $K$ whether $N=100$ or $N=1000$ is a better choice.
In particular, when $K\in\{5,10\}$ PL-Rank-3 with $N=100$ has the highest performance on MSLR and Istella and is only outperformed by PL-Rank-3 with $N=1000$ on Yahoo!
Conversely, when $K\in\{50,100\}$, $N=1000$ is the better choice for PL-Rank-3 on all datasets.
This suggests that gradient estimation for larger values of $K$ is more prone to variance and thus requires more samples for stable optimization.
Overall, PL-Rank-2 appears very affected by variance across datasets and $K$ values, we mostly attribute this to the small number of epochs it can complete in 200 minutes.
For example, on Istella with $K=100$, PL-Rank-2 completes less than eight epochs whereas PL-Rank-3 can complete 116 epochs.
Interestingly, StochasticRank with $N=100$ has stable performance that is sometimes comparable with PL-Rank-3 with $N=1000$ on the Yahoo!\ dataset.
This indicates that the sample-efficiency of StochasticRank is actually better than PL-Rank-3, however, due to its low computational costs, PL-Rank-3 still outperforms it on MSLR and Istella and when $K\in\{5,10,20\}$ on Yahoo!
We conclude that PL-Rank-3 provides a clear and substantial improvement over PL-Rank-2, and in most scenarios also outperforms StochasticRank.

\section{Conclusion}

We have introduced PL-Rank-3 an \ac{LTR} algorithm to estimate the gradient of a \ac{PL} ranking model with the same computational complexity as the best sorting algorithms.
PL-Rank-3 could enable future metric-based \ac{LTR} to be applicable to ranking lengths and item collection sizes of much larger scales than previously feasible.

\subsection*{Code and data}
To facilitate reproducibility, this work only made use of publicly available data and our experimental implementation is publicly available at \url{https://github.com/HarrieO/2022-SIGIR-plackett-luce}.

\begin{acks}
This work was partially supported by the Google Research Scholar Program and made use of the Dutch national e-infrastructure with the support of the SURF Cooperative using grant no.\ EINF-1748.
All content represents the opinion of the author, which is not necessarily shared or endorsed by their respective employers and/or sponsors.
\end{acks}

\balance
\bibliographystyle{ACM-Reference-Format}
\bibliography{references}

\end{document}